\documentclass{article}
\usepackage{ijcai26}

\usepackage{times}
\usepackage{soul}
\usepackage{url}
\usepackage[hidelinks]{hyperref}
\usepackage[utf8]{inputenc}
\usepackage[small]{caption}
\usepackage{graphicx}
\usepackage{amsmath}
\usepackage{amsthm}
\usepackage{booktabs}
\usepackage{algorithm}
\usepackage{algorithmic}
\usepackage[switch]{lineno}
\usepackage{amssymb}

\usepackage{multirow}
\DeclareMathOperator*{\argmax}{arg\,max}
\usepackage{subcaption}

\newcommand\scalemath[2]{\scalebox{#1}{\mbox{\ensuremath{\displaystyle #2}}}}

\title{EvoCSFL: Surrogate-Assisted Evolutionary Client Selection for Efficient and Robust Federated Learning}

\author{
$^1$Lin Qiang
\and
$^1$Sun Xiaoyan\textsuperscript{*}
\and
$^2$Hu Yao\textsuperscript{*}
\and
$^1$Fang Wei
\\
\affiliations
$^1$Jiangnan University\\
$^2$The Hong Kong Polytechnic University\\
\emails
6243115013@stu.jiangnan.edu.cn,
\{xysun78, fangwei\}@jiangnan.edu.cn,
echo-yao.hu@polyu.edu.hk
}

\begin{document}

\maketitle

\begin{abstract}
The heterogeneity of client data and systems makes it difficult to achieve satisfactory convergence speed and robustness in federated learning with random client selection. To address this issue, this paper proposes a surrogate-assisted client evolutionary selection framework for federated learning. In this framework, some typical client selection strategies are first used to generate candidate sets, and a metric function that integrates model performance, communication latency, and energy consumption is developed to formulate the client selection problem as a combinatorial optimization one. Subsequently, a surrogate model is constructed using the candidate selections and metric to efficiently approximate the performance of selected client subsets. An evolutionary algorithm is employed to search the combinatorial space of client selections, guided by the surrogate model to accelerate convergence. Experiments on MNIST, CIFAR10, CINIC10, and TinyImageNet demonstrate that the proposed algorithm achieves faster convergence, lower energy consumption, and improved robustness compared to existing methods.
\end{abstract}

\section{Introduction}

Federated learning (FL)~\cite{RN20}, proposed by Google, is a privacy-preserving framework where clients train local models on private data and iteratively share only model updates with a central server. The server aggregates these updates to improve a global model, which is then sent back to clients to avoid sharing raw data and leverage edge resources. However, there are significant challenges to the practical deployment of FL, stemming from the diversity among clients. First, the data possessed by different clients is usually non-independent and non-identically distributed (Non-IID). This disparity in data distribution leads to inconsistent model update directions among clients~\cite{RN21}, which severely affects the convergence speed and final performance of the global model~\cite{RN22}. Second, clients vary significantly in computing power, network bandwidth, memory capacity, and battery life. These differences result in great differences in the time and energy required for local training~\cite{RN23}, significantly prolonging the overall training cycle.

The random client selection strategy is simple, but it is known to be ineffective in identifying high-performance clients and weakens the performance of global aggregated models, which greatly impedes rapid model convergence and hinders the achievement of the desired final accuracy. This insufficiency is particularly pronounced in data and local system heterogeneity environments. Additionally, random selection may select resource-limited clients, known as the "straggler problem" ~\cite{RN25,RN24}, which require extended periods to complete local training and model uploads. This significantly reduces the efficiency of the training process and leads to inefficient allocation of time and computational resources. Accordingly, designing intelligent client selection algorithms in FL to overcome data and system heterogeneity is quite important.

Current client selection methods typically require real-time or historical status information of client devices, including local training loss, model update quality, training duration, and energy consumption. Clearly, accurately obtaining this information in real-time is computationally expensive in the dynamic, heterogeneous environment of FL. For example, FedRank~\cite{RN27} requires all clients to perform at least one round of local training to submit their status. In contrast, Oort~\cite{RN28} uses historical client status information, but its decision mechanism essentially calculates a sampling probability based on clients' historical performance to randomly sample clients. These methods fail to capture the combinatorial nature of FL. In particular, evaluating each candidate client subset in FL is time-consuming and resource-limited, making client selection a typical expensive optimization problem. For FL, client selection is a combinatorial optimization problem with an exponentially growing search space; simple random sampling or probabilistic selection cannot effectively explore the vast solution space and is prone to local optima.

Motivated by the above, we propose a surrogate-assisted evolutionary client selection framework for federated learning, named EvoCSFL(Evolutionary Client Selection for FL). EvoCSFL formulates client selection as a combinatorial optimization problem and solves it using surrogate-assisted evolutionary algorithms (SAEAs). Compared to traditional methods, EvoCSFL offers three key advantages: (1) it treats client selection as a multi-objective combinatorial optimization problem, capturing collective subset characteristics that existing approaches overlook; (2) the surrogate model enables low-cost, high-efficiency prediction of any selection’s future composite score, avoiding the computational and communication overhead of real-time status acquisition; and (3) it harnesses the global search capability of evolutionary algorithms to efficiently explore the exponentially large combinatorial space, identifying near-optimal client sets with high convergence speed and robustness in dynamic, heterogeneous settings.

\section{Background and Related Work}

\subsection{Client Selection in FL}

In federated learning, the client selection strategy is crucial to achieve high-performance global models. Existing methods can be categorized into three groups based on how device status information is obtained, i.e., random-based methods, history-based methods and probe-based methods.

\paragraph{Random-based Methods.}
This is the most typical approach, like FedAvg~\cite{RN20}, which selects clients uniformly at random, ensuring fairness and zero overhead but offering no adaptation to data or system heterogeneity.

\paragraph{History-based methods.} 
These algorithms leverage past client performance. AFL~\cite{RN69} uses historical local loss to set sampling probabilities. Oort~\cite{RN28} prioritizes clients with higher loss, shorter training time, and fewer prior selections. FedSAE~\cite{RN29} adjusts participation likelihoods dynamically using historical stats. FedMCCS~\cite{RN30} incorporates location, resources, and data characteristics. TiFL~\cite{RN31} clusters clients according to the training duration, and Favor~\cite{RN35} applies Q-learning to infer data distributions from model weights, boosting convergence and energy efficiency. These methods do not need extra computing cost but may struggle in highly changed settings.

\paragraph{Probe-based methods.}
These methods require capturing the real-time status of each client via preliminary training. FedRank~\cite{RN27} uses a one-cycle “early exit” probe and a pre-trained ranking model to score clients; FedMarl~\cite{RN37} employs multi-agent reinforcement learning with lightweight probe rounds to gather loss and performance data in each round. Clearly, these approaches need extra resources to achieve probing.

\subsection{Combinatorial Optimization in FL}

The client selection problem is inherently a combinatorial optimization task: selecting the best subset of clients to optimize a multi-objective function involving model performance, training latency, and energy consumption. As the number of clients grows, the search space expands exponentially, rendering exact solutions intractable and necessitating efficient approximation strategies.

Common approaches include greedy algorithms, dynamic programming, integer programming, and heuristics. FedCS~\cite{RN65} uses a greedy method based on pre-collected resource data, but is easy to fall into local optimum. Dynamic and integer programming become infeasible when the size of clients fast increases~\cite{RN40,RN41}. FedGCS~\cite{RN55} maps selection into a continuous space and applies gradient descent, yet still risks local optima. Another study~\cite{RN41} proposes an online heuristic to jointly optimize client count and bandwidth under energy constraints. Since such problem is NP-hard~\cite{RN67}, exact polynomial-time solutions are generally impossible, making heuristic methods the practical choice.

Evolutionary algorithms (EAs), inspired by natural selection, maintain a population of candidate solutions and iteratively improve them via evolutionary operators, e.g., selection, crossover, and mutation. They excel at navigating complex, high-dimensional spaces and offer tunable trade-offs between solution quality and computational cost. For example, FedCSGA~\cite{RN68} uses a genetic algorithm to maximize a weighted sum of client count and model accuracy under a time budget. However, it requires pre-training all clients to estimate accuracy, which incurs significant computational overhead and creates costly optimization subproblems.

\subsection{Surrogate-assisted Evolutionary Algorithms}

Evaluating the performance of each combination of selected clients in FL is very time consuming, and is especially challenging for using EAs to find the optimal solution, i.e., it is a typical expensive optimization problem. Such problems involve objective or constraint evaluations that demand significant time, resources, or physical experiments. Traditional EAs are often impractical for such problems due to their reliance on numerous costly function evaluations. SAEAs address this by using a cheap-to-evaluate surrogate model to approximate the true objective, drastically reducing the need for expensive evaluations and improving efficiency.
The basic workflow of SAEAs typically follows the iterative steps: (1) Generate an initial population and evaluate individuals using the true objective. (2) Train or update a surrogate model with evaluated data. (3) Apply evolutionary operators to generate offspring, whose fitness is estimated by the surrogate. (4) Select promising candidates via an infill criterion for true evaluation and augment the training set. (5) Repeat above steps until convergence, yielding an approximate optimum.

In constructing surrogate models, researchers have extensively explored various machine learning techniques to approximate expensive true objective functions. Commonly used surrogate models include Kriging interpolation~\cite{RN42}, radial basis function (RBF) models~\cite{RN43}, Gaussian process (GP) models~\cite{RN44}, support vector machines (SVMs)~\cite{RN45}, and ensembles of multiple surrogate models. These models are widely adopted due to their strong performance in approximating nonlinear functions and providing predictive uncertainty estimates. Despite the wide variety of available models, selecting the most suitable surrogate model for a specific optimization problem remains challenging and often relies on empirical experience rather than clear theoretical guidance. For instance, traditional surrogate models struggle to capture sequential dependencies or temporal correlations inherent in complex optimization problems with time-series or history-dependent characteristics.

SAEAs have been successfully applied to both single- and multi-objective optimization. In single objective settings, Yang et al.~\cite{RN46} use a surrogate to classify and select promising subpopulations for faster convergence. In multi-objective optimization,~\cite{RN47} employs surrogates for pairwise solution comparisons instead of direct fitness prediction,~\cite{RN48} integrates Pareto active learning with surrogate-assisted particle swarm optimization and a simulated evolution-inspired mutation strategy to enhance solution quality. Recent advances include decomposition-based local learning~\cite{RN49} and interpretable machine learning for acceleration~\cite{RN50}, further expanding SAEAs to address complex optimization challenges.However, the application of SAEAs to client selection in federated learning remains unexplored. Our work is the first to introduce SAEAs into this domain, proposing the EvoCSFL framework to efficiently solve the expensive multi-objective combinatorial optimization problem of client selection.

\begin{figure*}[t]
  \centering
  \includegraphics[width=0.9\textwidth]{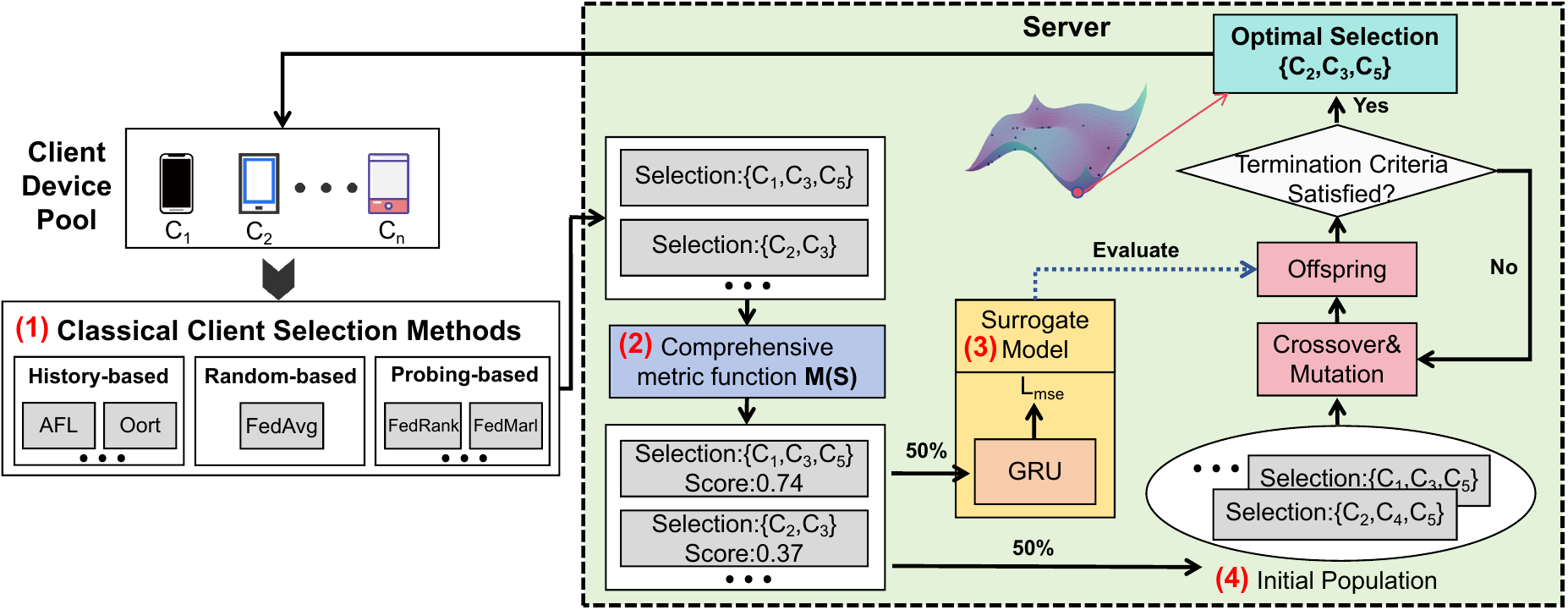}
  \caption{The EvoCSFL framework involves four main steps:  
(1) Collect good client selections using existing algorithms.  
(2) Score them each round with a comprehensive metric.  
(3) Train a surrogate model on half the data to predict selection performance.  
(4) Use remianing data to initialize and evolve toward the best selection.}
  \label{fig:framework}
\end{figure*}

\section{Framework of EvoCSFL}

The framework of the proposed EvoCSFL is demonstrated in Figure~\ref{fig:framework} and includes four main components: (1) obtaining candidate client selections; (2) calculating the evaluation metric on each candidate; (3) constructing a surrogate model based on the evaluated candidates; and (4) searching for optimal client selections using SAEAs. These components are detailed in the following subsections. 

\subsection{Problem Definition}

Consider a candidate pool of $U$ client devices $C = \{c_1, c_2, \dots, c_U\}$, where client $c_i$ possesses its own dataset $D_i$. A client selection is defined as $S = \{s_1, s_2, \dots, s_V\}$ where $S \subseteq C$. We guide client selection based on the following perspectives: (1) impact on the global model; (2) lower latency; (3) lower energy consumption. The corresponding comprehensive metric function $M(S)$ for evaluating the selected client solution $S$ is defined as follows:

\begin{equation}
\scalemath{0.8}{{M(S) = P_{\text{pref}}(S) \cdot \left( \frac{L}{P_L(S)} \right)^{\mathbb{I}(L < P_L(S)) \cdot \alpha} \cdot \left( \frac{E}{P_E(S)} \right)^{\mathbb{I}(E < P_E(S)) \cdot \beta}}}
\end{equation}

The formula consists of four main components:
\begin{itemize}
    \item $P_{pref}(S) = \frac{1}{|S|} \sum_{j \in S} \left( |D_j|  \sqrt{\frac{1}{|D_j|}\sum_{k \in D_j} \text{Loss}^2(k)} \right)$, representing the performance of $S$ on downstream tasks, where $D_j$ denotes the local training dataset of client j, and $\text{Loss}^2(k)$ represents the squared loss computed on data sample k.
    
    \item $P_{L}(S) = \max_{i \in S} \left( T_{i}^{train} + T_{i}^{comm} \right)$, representing the actual latency of $S$, which is determined by the maximum training and communication time among all clients in the selected subset. 
    
    \item $P_{E}(S) = \max_{i \in S} \left( E_{i}^{train} + E_{i}^{comm} \right)$, representing the actual energy consumption of $S$, defined as the maximum sum of local training and communication energy consumption across the selected clients.
    
    \item $L$ and $E$ denote the latency and energy consumption budgets, respectively. $\mathbb{I}(\cdot)$ is the indicator function that imposes a penalty when the actual resource usage exceeds these budgets.
\end{itemize}

For clients that have not participated in training, the aforementioned state information can only be obtained after they actually engage in training. Therefore, predicting the state of these devices during the pre-selection phase in a more cost-effective manner becomes a key challenge. Using surrogate models to replace actual evaluations can significantly reduce the computational resources required.

\subsection{Candidate Client Selections Generation}

We use three typical client selection methods discussed in subsection 2.1, i.e., TiFL, FedAvg and Oort, to generate initial client selections for the subsequent SAEAs process, i.e., surrogate modeling and evolutionary searching. Therefore, client selection sets together with performance evaluations are obtained here. With the above strategies, the dataset of selected clients during the $i$-th communication round is denoted as $D_h^i = \{(s_j,y_j)\}_{j=1}^N$. The performance evaluation $y_j$ for $s_j$ is calculated by normalizing the comprehensive metric function $M(S)$: 

\begin{equation}
y_j = \frac{M(s_j) - M_{\text{extmin}}}{M_{\text{extmax}} - M_{\text{extmin}}} \times \theta_s,
\end{equation}

\noindent , where $M(s_j)$ is the original score computed with the comprehensive metric function for the client selection $s_j$, $M_{\text{extmax}}$ and $M_{\text{extmin}}$ are the maximum and minimum values of $M(s_j)$ among all client selection sets in the current communication round. $\theta_s$ is a scaling factor used to map the score into the range $[0, \theta_s]$. The value of $y_j$ depends on the performance of the clients combination in its last training session. For clients that have never been selected in FL, the average performance of those selected clients is adopted here.

The final dataset $D_g^i$ obtained in communication round $i$ is constructed by combining $D_h^i$ with sampled historical data $D_h^{(1\sim i-1)}$, where $N_{\text{hist}}$ denotes the number of randomly selected samples. The specific formulas are given as follows:

\begin{equation}
D_h^{(1\sim i-1)} = \bigcup_{k=1}^{i-1} D_h^k
\end{equation}

\begin{equation}
D_g^i = D_h^i \cup \text{Sample}\left(D_h^{(1\sim i-1)}, N_{\text{hist}}\right)
\end{equation}

\subsection{Design of Surrogate Model}

In EvoCSFL, a surrogate model is designed to predict the performance of client selection strategies, guiding the subsequent evolutionary search. The model is trained on the historical dataset $ D_g $  to learn the mapping between client selection sequences and their observed performance. Federated learning is inherently dynamic client performance and states change continuously due to device network fluctuations. Importantly, the performance depends on the set of selected clients, not the order of selection. To enforce this order invariance, we augment the training data by randomly shuffling each client selection sequence multiple times.

At the model level, we employ a gated recurrent unit (GRU)~\cite{gru} as the encoder. The given sequence of a clients-combination $ S = \{s_1, s_2, \ldots, s_N\} $ is sequentially fed into the encoder, producing an output embedding sequence $ \mathbf{E}_S = [\mathbf{h}_1, \mathbf{h}_2, \ldots, \mathbf{h}_N] \in \mathbb{R}^{N \times d} $, where $ N $ is the length of the sequence $ S $, and $ d $ is the embedding dimension. We then compute the mean pooling over $ \mathbf{E}_S $ to obtain a unified embedding representation $ \bar{\mathbf{E}} = \frac{1}{N} \sum_{i=1}^N \mathbf{h}_i $ with $ \bar{\mathbf{E}} \in \mathbb{R}^d $. Finally, $ \bar{\mathbf{E}} $ is passed through a non-linear layer to estimate the score $ \hat{\mathbf{y}} $. 

To minimize the discrepancy between the estimated score and the true score, we adopt the mean squared error (MSE) loss, defined as:

\begin{equation}
L_{\text{mse}} = \text{MSE}(\mathbf{y}, \hat{\mathbf{y}}) = (\mathbf{y} - \hat{\mathbf{y}})^2.
\end{equation}

\begin{figure*}[!ht]
  \centering
  \includegraphics[width=0.9\textwidth]{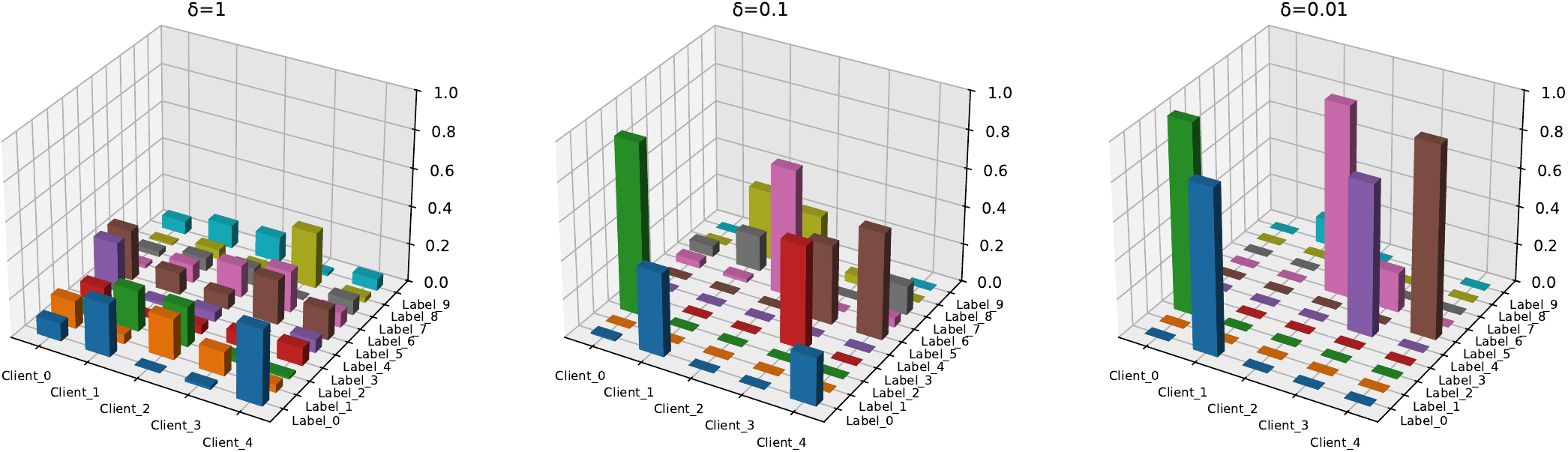}
  \caption{The distribution of labels under varying concentration levels, where lower concentration results in more extreme label distributions.}
  \label{fig:data_distribution}
\end{figure*}

\subsection{Searching by Genetic Algorithm}

Although the surrogate model maps client selections into a continuous space where gradient descent~\cite{RN55} could be applied, such local search methods are prone to getting trapped in suboptimal solutions due to their reliance on local gradient information and lack of global exploration. To address this, EvoCSFL employs a surrogate-assisted genetic algorithm (SAGA) for global combinatorial optimization and its procedure is illustrated in Algorithm~\ref{alg:evo_csfl}.

\begin{algorithm}[!ht]
\small
\caption{Procedure of SAGA}
\label{alg:evo_csfl}
\begin{algorithmic}[1]
\REQUIRE Heuristic-based Initialization $P_0 = \{S_1, S_2, \ldots, S_N\}$, where $S_i \subseteq C$, population size $N$, maximum iterations $T$, mutation rate $\mu$, surrogate model $f(\cdot)$, candidate clients $C = \{c_1, c_2, \ldots, c_U\}$
\STATE Initial Evaluation: $V_0 \leftarrow \{f(S_i) \mid S_i \in P_0\}$
\FOR{$t = 1$ to $T$}
    \STATE Selection: $P_t \sim \text{RouletteWheelSelect}(P_{t-1}, V_{t-1}, N)$
    \FOR{$j = 1$ to $N/2$}
        \STATE Crossover: $S'_{2j-1}, S'_{2j} \leftarrow \text{OX}(S_{r_1}, S_{r_2})$
        \STATE Mutation: For each $S' \in \{S'_{2j-1}, S'_{2j}\}$\\
        \quad$S' \leftarrow
        \begin{cases}
        \text{Swap}(S') & \text{with prob. } \mu \\
        \text{Replace}(S') & \text{with prob. } \mu/2
        \end{cases}$
        \STATE Evaluation: $v'_{2j-1} \leftarrow f(S'_{2j-1})$, $v'_{2j} \leftarrow f(S'_{2j})$
    \ENDFOR
    \STATE Elitist Preservation: $S_{\text{best}} \leftarrow \argmax\limits_{S \in P_{t-1}} f(S)$
    \STATE Update Population: $P_t \leftarrow P_{t-1} \cup \{S'_1, \ldots, S'_N\} \cup \{S_{\text{best}}\}$, 
    $V_t \leftarrow V_{t-1} \cup \{v'_1, \ldots, v'_N\} \cup \{f(S_{\text{best}})\}$
\ENDFOR
\STATE \textbf{Return} $S^* = \argmax\limits_{S \in P_T} f(S)$
\end{algorithmic}
\end{algorithm}

At the beginning, each client selection scheme is encoded as an individual using an integer sequence representation, where a sequence of unique client IDs denotes the selected subset. Given the exponential $2^{|C|}$ search space of possible combinations, random population initialization is inefficient, risking slow convergence or failure. To resolve this, the algorithm starts evolutionary iterations from a heuristic-based initial population built with high-quality historical selection sets stored in $ D_g $. Such an initial population focuses search in the most promising subspaces.

In the evolutionary loop, a GRU-based surrogate model $f(\cdot)$ serves as a low-cost fitness evaluator. It quickly assesses each potential client combination in the population by predicting its comprehensive score, enabling the GA to efficiently explore the large solution space without costly training evaluations.

For selection, we use roulette wheel selection to calculate each individual’s fitness proportion relative to the total fitness sum. This approach retains high fitness individuals while still allowing lower fitness ones to be chosen, preserving population diversity. For crossover, we replace conventional Single-Point Crossover (SPX)~\cite{crossover} with Ordered Crossover (OX)~\cite{RN56} as the primary operator. OX preserves a contiguous segment from one parent and fills remaining positions of the other parent’s order, ensuring no duplicate clients IDs in offspring and avoid conflict detection.

To boost the algorithm’s exploration ability, we design a hybrid mutation strategy: with probability $\mu$, swap mutation exchanges two clients’ positions within an individual; with probability $\mu/2$, replacement mutation swaps an existing client for a random candidate pool member. This dual mechanism balances local search precision and population diversity. Finally, an elite preservation strategy is implemented.

\begin{table*}[ht]
\centering
\resizebox{\textwidth}{!}{
\begin{tabular}{lc|ccc|ccc|ccc|ccc}
\hline
\multicolumn{2}{c|}{\multirow{2}{*}{\textbf{Dataset \& Model}}} & \multicolumn{3}{c|}{\textbf{MNIST-LeNet5}} & \multicolumn{3}{c|}{\textbf{CIFAR10-ResNet18}} & \multicolumn{3}{c|}{\textbf{CINIC10-VGG16}} & \multicolumn{3}{c}{\textbf{TinyImageNet-ShuffleNet}} \\
\cline{3-14}
& & Acc(\%)$\uparrow$ & Energy$\downarrow$ & Speed$\uparrow$ & Acc(\%)$\uparrow$ & Energy$\downarrow$ & Speed$\uparrow$ & Acc(\%)$\uparrow$ & Energy$\downarrow$ & Speed$\uparrow$ & Acc(\%)$\uparrow$ & Energy$\downarrow$ & Speed$\uparrow$ \\
\hline
\multirow{7}{*}{\rotatebox[origin=c]{90}{\textbf{IID}}}
& FedAvg   & 98.75   & 100\%  & 1$\times$ & 87.30  & 100\%  & 1$\times$ & 66.01  & 100\%  & 1$\times$ & 41.27  & 100\%  & 1$\times$ \\
& AFL      & 98.80   & 61.3\% & 1.62$\times$ & 86.85  & 74.4\% & 1.32$\times$ & 66.75  & 74.3\% & 1.31$\times$ & 42.66  & 75.8\% & 1.29$\times$ \\
& TiFL     & 98.86   & 60.0\% & 1.72$\times$ & 87.28  & 49.5\% & 1.99$\times$ & 67.41  & 46.1\% & 2.08$\times$ & 41.96  & 60.6\% & 1.62$\times$ \\
& Oort     & 98.97   & 44.2\% & 2.12$\times$ & 86.68  & 43.3\% & 2.08$\times$ & 67.38  & 48.0\% & 1.87$\times$ & 43.26  & 68.1\% & 1.42$\times$ \\
& FedRank  & ${99_{(32)}}^{1}$  & \textbf{40.1\%} & 1.56$\times$ & 87.69  & 44.4\% & 1.67$\times$ & 68.58  & 52.9\% & 2.01$\times$ & 43.54  & 63.3\% & 1.81$\times$ \\
& FedGCS   & $\mathbf{{99_{(28)}}^{1}}$  & -      & -           & 88.42  & -      & -           & \textbf{69.09}  & -      & -           & 43.35  & -      & - \\
& EvoCSFL  & ${99_{(44)}}^{1}$  & 41.4\% & \textbf{2.28$\times$} & \textbf{88.72}  & \textbf{35.8\%} & \textbf{2.49$\times$} & 68.76  & \textbf{31.5\%} & \textbf{2.70$\times$} & \textbf{44.68}  & \textbf{48.0\%} & \textbf{1.87$\times$} \\
\hline
\multirow{7}{*}{\rotatebox[origin=c]{90}{\textbf{Non-IID$^{2}$}}}
& FedAvg   & 87.61   & 100\%  & 1$\times$ & 52.93  & 100\%  & 1$\times$ & 39.07  & 100\%  & 1$\times$ & 26.65  & 100\%  & 1$\times$ \\
& AFL      & 90.23   & 63.5\% & 1.53$\times$ & 50.78  & 62.0\% & 1.51$\times$ & 40.88  & 65.6\% & 1.45$\times$ & 29.00  & 64.4\% & 1.48$\times$ \\
& TiFL     & 91.59   & 58.7\% & 1.77$\times$ & 56.15  & 52.8\% & 1.88$\times$ & 42.10  & 51.7\% & 1.88$\times$ & 28.43  & 64.5\% & 1.54$\times$ \\
& Oort     & 93.18   & 51.4\% & 1.84$\times$ & 61.02  & 46.1\% & 1.95$\times$ & 45.15  & 46.1\% & 1.95$\times$ & 29.07  & 47.1\% & 1.92$\times$ \\
& FedRank  & 93.67   & 47.4\% & 1.48$\times$ & 59.11  & 50.8\% & 1.78$\times$ & 47.07  & 55.2\% & 1.75$\times$ & 30.04  & 67.4\% & 1.83$\times$ \\
& FedGCS   & 95.83   & -      & -           & 61.62  & -      & -           & 46.27  & -      & -           & 29.51  & -      & - \\
& EvoCSFL  & \textbf{96.52}   & \textbf{45.5\%} & \textbf{1.92$\times$} & \textbf{63.31}  & \textbf{38.0\%} & \textbf{2.37$\times$} & \textbf{48.25}  & \textbf{33.9\%} & \textbf{2.51$\times$} & \textbf{30.83}  & \textbf{40.6\%} & \textbf{2.23$\times$} \\
\hline
\end{tabular}
}
\caption{Comparison of EvoCSFL with other client selection methods, including energy consumption and training speed metrics. \textsuperscript{1} Training is terminated early upon reaching a target accuracy(99\%). \textsuperscript{2} For the MNIST dataset, $\delta = 0.01$, for other dataset, $\delta = 0.1$.}
\label{tab:comparison}
\end{table*}

\section{Experiment}

\subsection{Experiment Setup}

\paragraph{Simulation setup.}

We use the Plato framework~\cite{plato}, an open-source FL platform built on PyTorch, to implement the experimental simulation. On this platform, an FL system with 100 mobile clients and a central model aggregation server is considered. To capture heterogeneous client training time, a Pareto distribution with shape parameter $\gamma = 1$, denoted as $p_k \sim \text{Pareto}(\gamma = 1)$ is applied to simulate the per-client computation time. For energy consumption, we assume a normal distribution with mean 1 and standard deviation 0.2 to describe the per-minute energy usage (in joules). To ensure realistic values, all generated samples are clipped to lie within the interval $[0.5, 1.5]$.

\paragraph{Datasets and models.}
We adopt typical models and datasets from the computer vision domain to evaluate the proposed algorithm, including: LeNet5~\cite{lenet5} trained on MNIST~\cite{mnist},
ResNet18~\cite{resnet18} trained on CIFAR10~\cite{cifar10},
VGG16~\cite{vgg16} trained on CINIC10~\cite{cinic10},
ShuffleNet~\cite{shufflenet} trained on TinyImageNet~\cite{tinyimagenet}.
Each dataset contains 10 classes, and the classification difficulty increases progressively across these datasets. To simulate data heterogeneity, we model the local data label distribution at each client using a Dirichlet distribution. For a $K$-class classification task, each client’s label distribution $\boldsymbol{p}_i = (p_{i,1}, p_{i,2}, \ldots, p_{i,K})$ is sampled from $\boldsymbol{p}_i \sim \text{Dirichlet}(\delta \cdot \mathbf{1}_K)$, where $\mathbf{1}_K$ is a $K$-dimensional vector of ones and the concentration parameter $\delta$ controls the level of data heterogeneity, as shown in Figure~\ref{fig:data_distribution}.

\paragraph{Baselines.}
We compare EvoCSFL with six state-of-the-art client selection methods: FedAvg~\cite{RN20}, AFL~\cite{RN69}, TiFL~\cite{RN31}, Oort~\cite{RN28}, FedRank~\cite{RN27}, and FedGcs~\cite{RN55}. We use Time to Accuracy (ToA), a metric that measures how long it takes for federated learning to reach a target accuracy. Our experiments aim to demonstrate that EvoCSFL:
(1) converges faster under both IID and Non-IID data;
(2) reduces energy use and minimizes straggler effects in heterogeneous systems; and
(3) remains robust in dynamic, resource-limited federated settings thanks to its surrogate-assisted evolutionary search.

\paragraph{Parameter Settings.}
At the beginning of each communication round, $K=10$ clients are selected from the device pool of size $N=100$. Each client performs 5 epochs of training with a learning rate of 0.01 and batch size of 32. For dataset $D_g$, we generate 100 distinct client-combinations for each of the three methods, resulting in 300 candidates in total. The scaling factor $\theta_s$ is set to 0.95, with $\alpha = 1$, $\beta = 1$, and $N_{\text{hist}} = 200$. To simulate order invariance, we randomly shuffle the order of each group of data twice. For the surrogate model, the hidden layer size is set to 64, trained for 100 epochs with batch size 32 and learning rate 0.001. The remaining 50\% of the data are used to initialize the heuristic population of size 300 with 150 evolutionary iterations and mutation probability $\mu = 0.1$ to ensure the evolution diversity.

\begin{figure}[ht]
\centering
\includegraphics[width=0.8\linewidth]{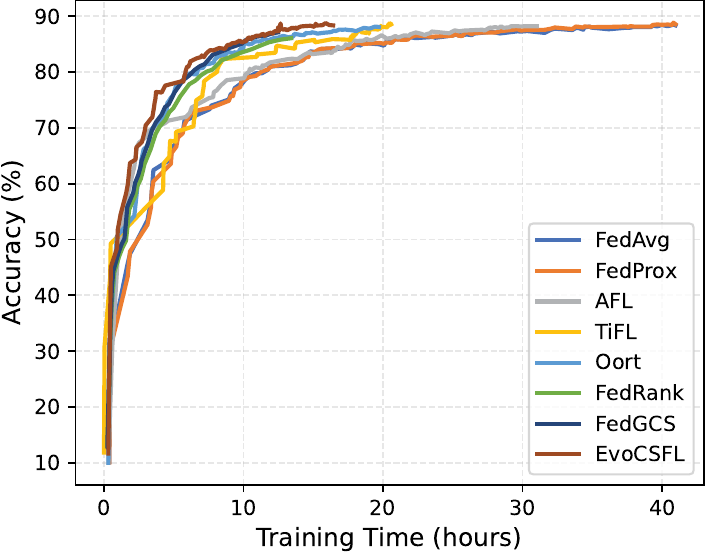}
\caption{Performance on the ToA metric when training ResNet18 on CIFAR10 (IID) dataset.}
\label{fig:cifar10_iid}
\end{figure}

\subsection{Overall Performance}

Table~\ref{tab:comparison} presents prediction accuracy, system energy consumption and training speed under IID and Non-IID configurations. Compared with FedGCS adopting gradient-based continuous optimization prone to local optima, EvoCSFL obtains superior performance via surrogate-assisted global evolutionary searching. In IID scenarios, EvoCSFL achieves the fastest convergence on all four benchmarks, reaching 2.49× and 2.70× speedup against FedAvg on CIFAR10-ResNet18 and CINIC10-VGG16 respectively. Such acceleration originates from carefully selecting high-quality clients and mitigating straggler overhead. As plotted in Figure~\ref{fig:cifar10_iid}, EvoCSFL exhibits a far steeper accuracy ascent than all baselines including FedGCS, verifying that the surrogate model combined with evolutionary optimization rapidly identifies optimal client subsets to expedite convergence. Meanwhile, its energy cost is lowered to only 35.8\% and 31.5\% of FedAvg on the above two datasets.
EvoCSFL gains an average IID accuracy improvement of 1.95\% over FedAvg. In challenging Non-IID environments, it outperforms all competitors across all datasets: the average accuracy rises by 7.5\% against FedAvg and surpasses the suboptimal FedGCS by 1.69\% on CIFAR10. These observations validate that the surrogate model exploits historical client statistics for precise performance prediction, supporting adaptive, robust client selection to cope with data heterogeneity and stabilize global training convergence.

\begin{figure*}[ht]
\centering
  \begin{subfigure}[b]{0.3\linewidth}
    \centering
    \includegraphics[width=1\linewidth]{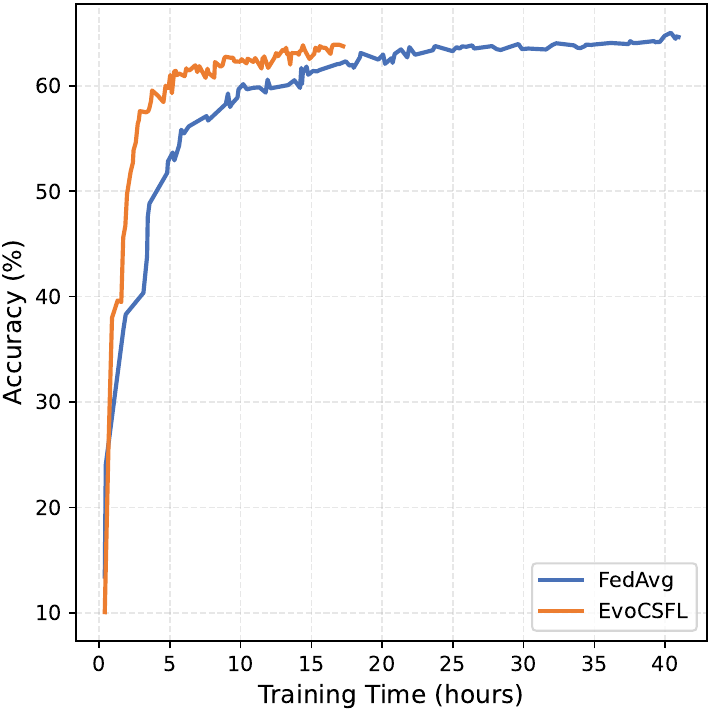}
    \caption{CIFAR10-ResNet18 ($\delta = 1$)}
    \label{fig:different_concentration_1}
  \end{subfigure}
  \hfill
  \begin{subfigure}[b]{0.3\linewidth}
    \centering
    \includegraphics[width=1\linewidth]{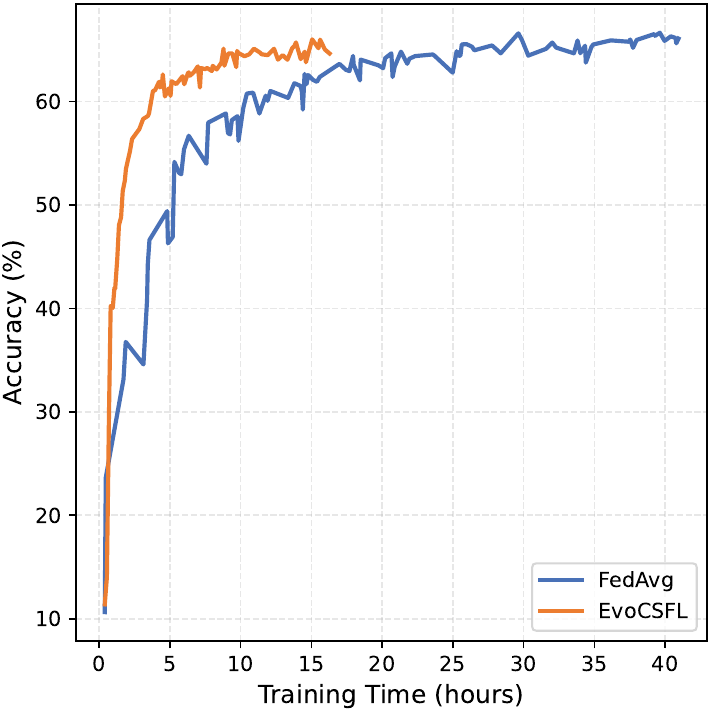}
    \caption{CIFAR10-ResNet18 ($\delta = 0.5$)}
    \label{fig:different_concentration_05}
  \end{subfigure}
  \hfill
  \begin{subfigure}[b]{0.3\linewidth}
    \centering
    \includegraphics[width=1\linewidth]{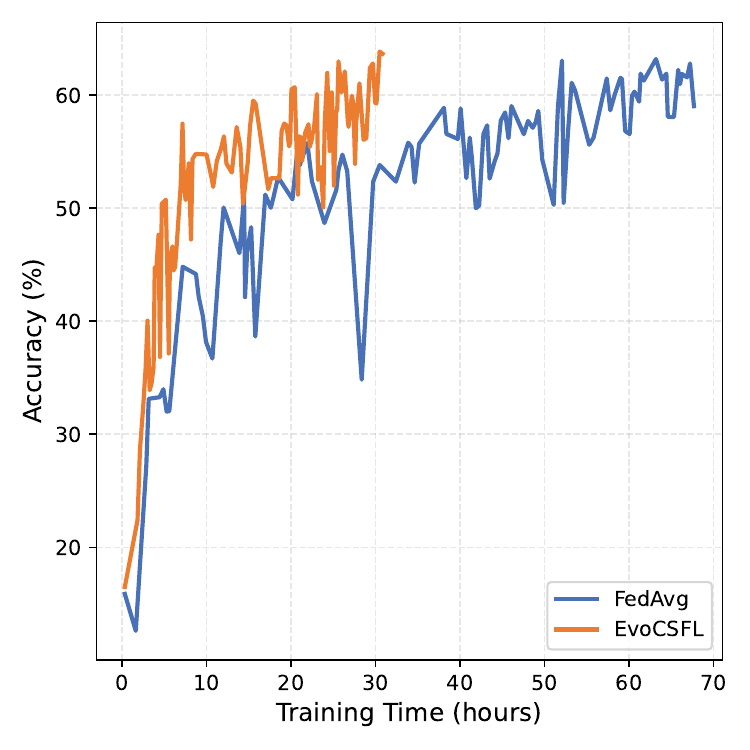}
    \caption{CIFAR10-ResNet18 ($\delta = 0.1$)}
    \label{fig:different_concentration_01}
  \end{subfigure}
\caption{ToA performance under different levels of data heterogeneity.}
\label{fig:different_concentration}
\end{figure*}

\subsection{Performance on Different Heterogeneous setup}
Figure~\ref{fig:different_concentration} compares the ToA performance of EvoCSFL and FedAvg on CIFAR10-ResNet18 under varying data heterogeneity controlled by $\delta$. EvoCSFL consistently achieves higher training efficiency and robustness across all settings. Under low ($\delta=1$) and moderate ($\delta=0.5$) heterogeneity (Figures~\ref{fig:different_concentration_1} and~\ref{fig:different_concentration_05}), it converges significantly faster, with steeper early accuracy gains and sustained superiority, leading to shorter time-to-target accuracy. Even under extreme heterogeneity ($\delta=0.1$, Figure~\ref{fig:different_concentration_01}), where both methods suffer high fluctuations due to severe label imbalance, EvoCSFL exhibits smaller variance and a clearer upward trend, demonstrating greater stability. This underscores its robustness: the surrogate model leverages historical client behavior to predict combination performance, enabling adaptive, informed selection that mitigates Non-IID challenges and supports stable global convergence.

\subsection{Analysis of Important Operators}

\begin{figure*}[t]
\centering
\parbox{0.32\textwidth}{
    \centering
    \includegraphics[width=\linewidth]{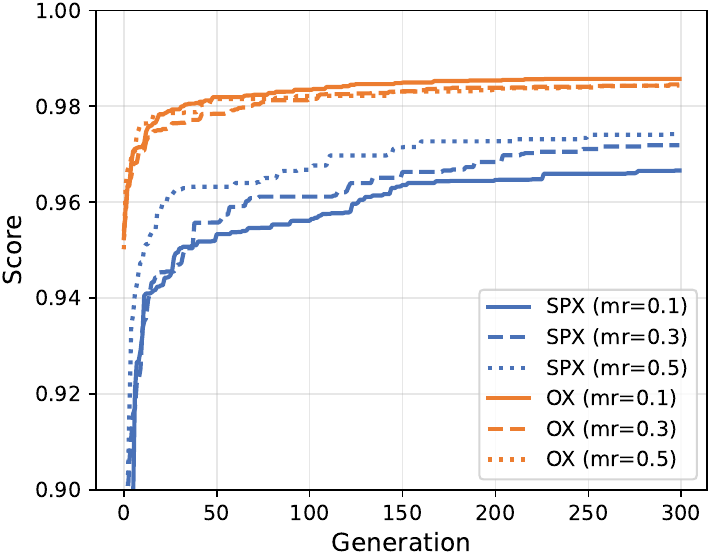}\\
    \vspace{0.5em}
    \caption{Comparison of convergence performance between SPX and OX crossover operators.}
    \label{fig:generation}
}
\hfill
\parbox{0.32\textwidth}{
    \centering
    \includegraphics[width=\linewidth]{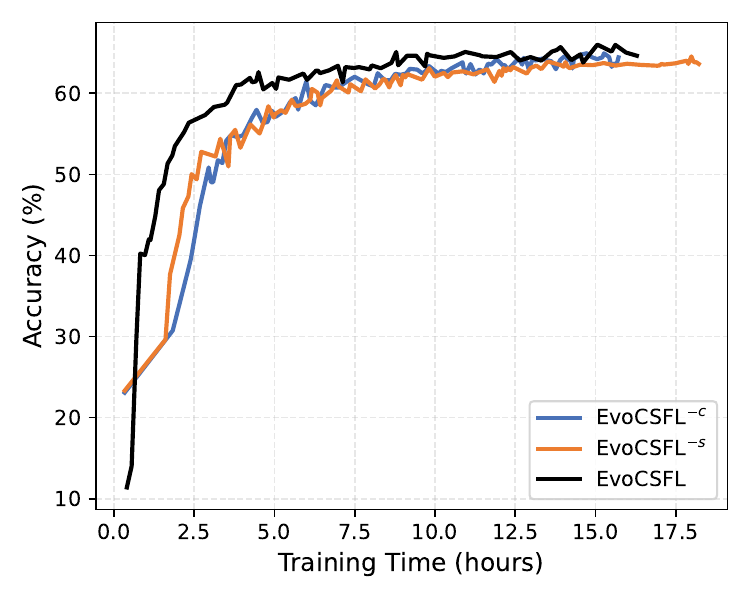}\\
    \vspace{0.5em}
    \caption{Ablation study of EvoCSFL, comparing its variants to a ResNet18 model trained on CIFAR10 with $\delta = 0.5$.}
    \label{fig:ablation}
}
\hfill
\parbox{0.32\textwidth}{
    \centering
    \includegraphics[width=\linewidth]{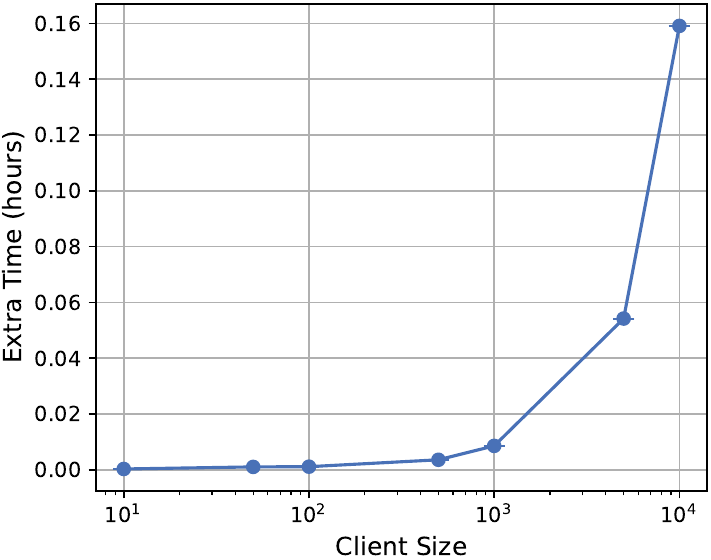}\\
    \vspace{0.5em}
    \caption{Average extra time overhead of EvoCSFL under different client scales, with a fixed client selection rate of 10\%.}
    \label{fig:extra_cost}
}
\end{figure*}

\paragraph{Convergence of crossover operators.}
As shown in Figure~\ref{fig:generation}, we compare crossover operators under varying mutation rates via repeated trials. The OX operator stands out by ensuring that offspring do not have duplicate clients, thereby avoiding costly conflict checks. In contrast, the SPX operator often generates invalid offspring with duplicates due to random segment swapping, severely slowing convergence. OX converges in approximately 150 rounds, whereas SPX fails to converge even after 300 rounds. A heuristic initialization strategy quickly directs the search toward high-quality regions, reducing the search space and enabling strong performance without high mutation rates.

\paragraph{Ablation study.}
We further illustrate the effectiveness of key components of EvoCSFL by removing corresponding modules. Two variants are tested: (1) EvoCSFL$^{-c}$ with only random-based methods to construct training data $D_g$; (2) EvoCSFL$^{-s}$, which removes the surrogate model and uses a greedy strategy based solely on the current composite fitness, i.e.,  $\arg\max_{S \in D_g} M(S)$.

As shown in Figure~\ref{fig:ablation}, EvoCSFL outperforms both variants. The improvement over EvoCSFL$^{-c}$, which omits the heuristic-based population initialization, underscores the effectiveness of this design in constructing a high-quality initial population and guiding the evolutionary search toward superior solutions. EvoCSFL$^{-s}$’s inferior performance shows that greedy selection fails to capture historical client behavior or adapt to dynamics. In contrast, the surrogate model leverages historical data to predict selection performance efficiently, enabling robust, adaptive decisions under client-side uncertainty.

\paragraph{Sensitivity Analysis of $\alpha$ and $\beta$}

\begin{table}[t]
\centering
\small
\setlength{\tabcolsep}{4pt}
\renewcommand{\arraystretch}{1.05}
\caption{Sensitivity of EvoCSFL to $\alpha$ and $\beta$ on
CIFAR10--ResNet18 (Non-IID, $\delta\!=\!0.1$). $\beta\!=\!1$ when varying
$\alpha$; $\alpha\!=\!1$ when varying $\beta$.}
\label{tab:alpha_beta}
\resizebox{0.49\textwidth}{!}{
\begin{tabular}{c|ccc|ccc}
\toprule
\multirow{2}{*}{Value} &
\multicolumn{3}{c|}{Vary $\alpha$ ($\beta\!=\!1$)} &
\multicolumn{3}{c}{Vary $\beta$ ($\alpha\!=\!1$)} \\
 & Acc(\%)$\uparrow$ & Energy$\downarrow$ & Speed$\uparrow$
 & Acc(\%)$\uparrow$ & Energy$\downarrow$ & Speed$\uparrow$ \\
\midrule
0    & 62.18 & 44.9\% & 1.87$\times$ & 62.55 & 43.5\% & 2.08$\times$ \\
0.5  & 62.94 & 39.1\% & 2.05$\times$ & 63.02 & 47.5\% & 2.15$\times$ \\
1    & \textbf{63.31} & 38.0\% & 2.37$\times$ & \textbf{63.31} & 38.0\% & \textbf{2.37$\times$} \\
2    & 63.05 & \textbf{34.7\%} & 2.41$\times$ & 62.88 & 33.1\% & 2.25$\times$ \\
4    & 60.42 & 37.1\% & \textbf{2.68$\times$} & 60.07 & \textbf{32.4\%} & 2.09$\times$ \\
\bottomrule
\end{tabular}
}
\end{table}

We analyze hyperparameters \(\alpha\) and \(\beta\) as show in table~\ref{tab:alpha_beta}. \(\alpha\) controls the penalty weight for clients exceeding latency threshold L, and \(\beta\) penalizes devices surpassing energy limit E. Larger \(\alpha\) or \(\beta\) introduces heavier punishment for clients violating corresponding constraints, while the values of L and E are determined by practical engineering requirements on maximum tolerable training time and device energy consumption.
With fixed \(\beta=1\), increasing \(\alpha\) consistently improves overall training speed owing to stricter latency punishment against stragglers. Accuracy peaks at \(\alpha=1\) (63.31\%), whereas an over-large \(\alpha\) excessively excludes high-quality clients and deteriorates model accuracy despite further speed gains.When fixing \(\alpha=1\), the rise of \(\beta\) continuously reduces system energy consumption via enhanced penalty on high-power clients. Optimal accuracy is obtained at \(\beta=1\), and overly enlarged \(\beta\) sacrifices classification performance to pursue lower energy usage.In practical engineering, FL requires balanced optimization on training time L and device energy E. The setting \(\alpha=\beta=1\) strikes the optimal trade-off among accuracy, speed and energy cost.

\begin{table}[t]
\centering
\small
\setlength{\tabcolsep}{2pt}
\begin{tabular}{@{}ccccc@{}}
\toprule
\multirow{2}{*}{\textbf{Model}} & \textbf{MSE} & \textbf{MAE} & $\mathbf{R^2}$ & \multirow{2}{*}{\textbf{Time (ms)}} \\
& \textbf{(mean±std)} & \textbf{(mean±std)} & \textbf{(mean±std)} & \\
\midrule
RBF  & 0.0267±0.0038 & 0.1235±0.0091 & 0.6356±0.0506 & \textbf{0.054} \\
GRU  & \textbf{0.0103±0.0012} & \textbf{0.0756±0.0040} & \textbf{0.8577±0.0163} & 0.716 \\
LSTM & 0.0121±0.0014 & 0.0820±0.0045 & 0.8312±0.0207 & 0.796 \\
\bottomrule
\end{tabular}
\caption{Comparison of Different Surrogate Models}
\label{tab:surrogate_models}
\end{table}

\paragraph{Surrogate model fitting ability.}
Table~\ref{tab:surrogate_models} shows the fitting ability of different surrogate models using Mean Squared Error (MSE), Mean Absolute Error (MAE), coefficient of determination ($\mathbf{R^2}$) and single inference latency. It can be concluded that the GRU model excels in both accuracy and stability with the smallest standard deviations, demonstrating strong consistency across trials. Long short-term memory (LSTM)~\cite{lstm} model performs adequately but falls slightly behind. Despite its fast inference, the RBF model shows poor fitting capability and significant performance variability.

\subsection{Extra Cost Analysis}
To evaluate the system overhead, Figure~\ref{fig:extra_cost} examines the time and memory costs of EvoCSFL. Experiments are conducted on the cinic10 by training a VGG16 model on an NVIDIA RTX 3090 GPU and an Intel Core i5 14600KF CPU. With 100 clients, EvoCSFL introduces an average additional runtime of only 3.98 seconds per communication round, which corresponds to just 70\% of the average training time of 9.2 minutes per round, satisfying stringent performance requirements. Memory consumption is primarily determined by the model itself, with VGG16 occupying 56.25 MB. In contrast, EvoCSFL incurs a negligible memory footprint of only 0.667 MB even when scaled to 10,000 clients, amounting to 1.2\% of the total memory overhead.

\section{Conclusion}
We present EvoCSFL, a surrogate-assisted evolutionary client selection framework to enhance federated learning with system and data heterogeneity. EvoCSFL formulates client selection as a combinatorial optimization problem and uses a surrogate model to efficiently evaluate candidate client subsets. Furthermore, EvoCSFL employs evolutionary algorithm to explore the solution space. Our framework contributes to federated learning by demonstrating that client selection can be treated as a dynamic combinatorial optimization task. Extensive experiments show that EvoCSFL outperforms existing methods with minimal computational overhead. Its application in practical problems will be further studied.

\bibliographystyle{named}
\bibliography{EvoCSFL}

\end{document}